\documentclass[lettersize,journal]{IEEEtran}
\usepackage{amsmath,amsfonts}
\usepackage{array}
\usepackage{textcomp}
\usepackage{stfloats}
\usepackage{url}
\usepackage{verbatim}
\usepackage{graphicx}
\usepackage{cite}

\usepackage{bm}
\usepackage{multirow}
\usepackage{caption}
\usepackage{subcaption}
\usepackage[linesnumbered,ruled,vlined]{algorithm2e}
\usepackage{amssymb} 
\usepackage{colortbl}
\usepackage{pifont}
\usepackage{booktabs}
\usepackage{rotating}
\usepackage{makecell}
\usepackage{titletoc}
\usepackage{listings}
\usepackage{xcolor} 
\lstdefinestyle{py}{
  language=Python,
  basicstyle=\ttfamily\footnotesize, 
  frame=none,                        
  numbers=none,
  showstringspaces=false,
  columns=fullflexible,
  keepspaces=true,
  breaklines=true,
  keywordstyle=\color{blue},
  commentstyle=\color{gray},
  stringstyle=\color{teal},
}
\usepackage{cleveref}
\crefname{table}{Table}{Tables}
\crefname{figure}{Figure}{Figures}
\crefname{algorithm}{Algorithm}{Algorithms}
\crefname{section}{Section}{Sections}
\crefname{appendix}{Appendix}{Appendices}
\crefname{equation}{Eq.}{Eqs.}

\definecolor{lightblue}{rgb}{0.21,0.49,0.74}
\definecolor{soft_red}{RGB}{240, 110, 110}
\definecolor{soft_blue}{RGB}{100, 170, 220}

\begin{document}

\title{Learn Faster and Remember More: Balancing Exploration and Exploitation for Continual Test-time Adaptation}

\author{Pinci Yang, Peisong Wen, Ke Ma, and Qianqian Xu\IEEEauthorrefmark{1},~\IEEEmembership{Senior Member,~IEEE}
\IEEEcompsocitemizethanks{
\IEEEcompsocthanksitem Pinci Yang and Ke Ma are with the School of Electronic, Electrical and Communication Engineering, University of Chinese Academy of Sciences, Beijing 100049, China (email: \texttt{yangpinci23@mails.ucas.ac.cn; make@ucas.ac.cn}).\protect\\
\IEEEcompsocthanksitem Peisong Wen is with the School of Computer Science and Technology, University of Chinese Academy of Sciences, Beijing 101408, China (email: \texttt{wenpeisong@ucas.ac.cn}).\protect\\
\IEEEcompsocthanksitem Qianqian Xu is with the Key Laboratory of Intelligent Information Processing, Institute of Computing Technology, Chinese Academy of Sciences, Beijing 100190, China (email: \texttt{xuqianqian@ict.ac.cn}).\protect\\
}
\thanks{Corresponding authors: Qianqian Xu.}
}

\markboth{Journal of \LaTeX\ Class Files,~Vol.~14, No.~8, August~2021}%
{Shell \MakeLowercase{\textit{et al.}}: A Sample Article Using IEEEtran.cls for IEEE Journals}


\maketitle

\begin{abstract}
Continual Test-Time Adaptation~(CTTA) aims to adapt a source pre-trained model to continually changing target domains during inference. 
As a fundamental principle, an ideal CTTA method should rapidly adapt to new domains (exploration) while retaining and exploiting knowledge from previously encountered domains to handle similar domains in the future.
Despite significant advances, balancing exploration and exploitation in CTTA is still challenging: 
\textbf{1)} Existing methods focus on adjusting predictions based on deep-layer outputs of neural networks. However, domain shifts typically affect shallow features, which are inefficient to be adjusted from deep predictions, leading to dilatory exploration;
\textbf{2)} A single model inevitably forgets knowledge of previous domains during the exploration, making it incapable of exploiting historical knowledge to handle similar future domains.
To address these challenges, this paper proposes a mean teacher framework that strikes an appropriate \textit{\textbf{B}alance between \textbf{E}xploration and \textbf{E}xploitation} (\textbf{\textit{BEE}}) during the CTTA process. For the former challenge, we introduce a Multi-level Consistency Regularization (MCR) loss that aligns the intermediate features of the student and teacher models, accelerating adaptation to the current domain. For the latter challenge, we employ a Complementary Anchor Replay (CAR) mechanism to reuse historical checkpoints (anchors), recovering complementary knowledge for diverse domains.
Experiments show that our method significantly outperforms state-of-the-art methods on several benchmarks, demonstrating its effectiveness for CTTA tasks.
\end{abstract}


\section{Introduction} \label{sec:intro}
\begin{figure}[t]
    \centering
    \includegraphics[width=0.45\textwidth]{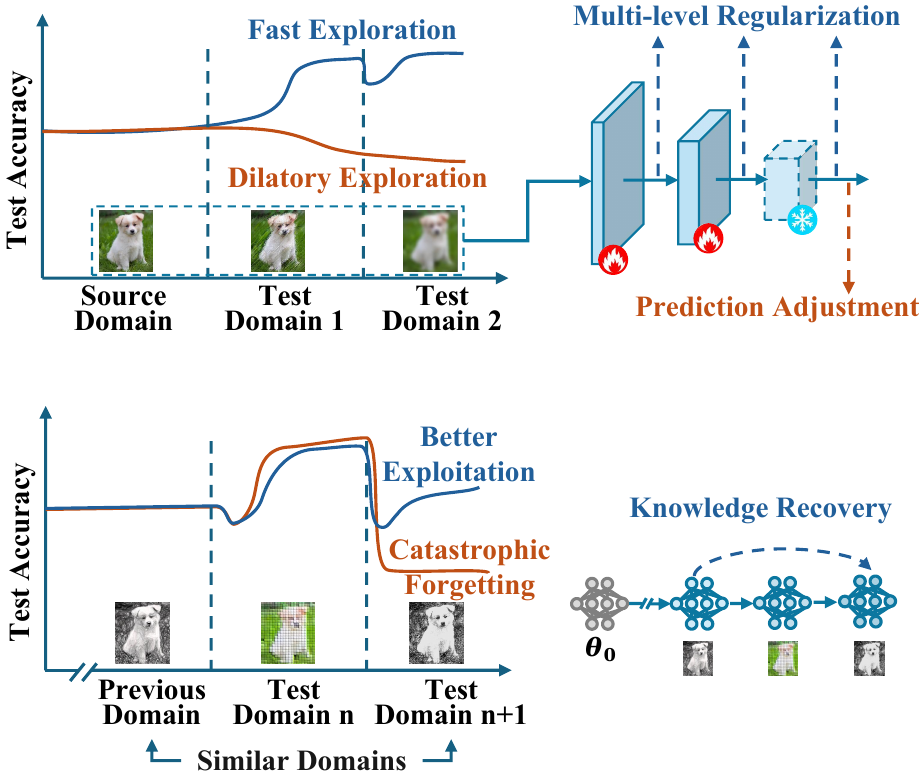}
    \caption{Our goal is to make an exploration-exploitation balance across continually shifting domains. Top: Multi-level regularization achieves fast exploration in the current domain, while prediction adjustment fails to update shallow parameters effectively. Bottom: Knowledge recovery benefits exploitation of previous domains, thereby preserving performance when domain shifts.}
    \label{fig:motivation}
\end{figure}

\IEEEPARstart{A}{} basic assumption of traditional machine learning is the independent and identical distribution of training and test data \cite{wang2022generalizing, zhou2022domain, liu2021towards}. However, this assumption is often violated in real-world scenarios due to non-stationary environments. In such cases, the test domain (target) deviates from the training domain (source), leading to weak generalization and performance degradation. For example, in autonomous vehicles \cite{janai2020computer}, the vision sensors might fail under varying weather and location. To mitigate the challenges posed by dynamically shifting distributions, \textit{\textbf{C}ontinual \textbf{T}est-\textbf{T}ime \textbf{A}daptation~(\textbf{CTTA})} \cite{wang2022continual} aims to adapt the pre-trained source model to various target domains while inferring test data. 
The fundamental target of CTTA lies in extracting beneficial information from the following domains while resisting adverse effects:
\begin{itemize}
    \item \textbf{Source domain:} The data domain used during the model's pretraining phase, which is significantly more diverse than the test data;
    \item \textbf{Previous test domains:} Data domains the model has previously adapted to during the testing phase, some of which contain complementary information to the current domain;
    \item \textbf{Current domain:} The data domain is currently awaiting adaptation and inference, with the smallest amount of data.
\end{itemize}

As illustrated in \cref{fig:motivation}, on the one hand, an ideal model should \textbf{rapidly explore the current domain} to maintain satisfactory performance before a domain shift occurs. On the other hand, the model should \textbf{exploit the source and previous domains} to prevent overfitting to the current domain, which could otherwise lead to significant performance degradation following domain changes.

To explore the current domain without ground truth labels, early studies \cite{wang2020tent, jang2023test} reduce prediction uncertainty by minimizing entropy in test batches. However, due to the limited exploitation of historical information, these methods are prone to overfitting to noisy pseudo-labels in the current domain. This accumulation of errors ultimately leads to mode collapse \cite{niu2023towards, hoang2024persistent}.
To address this issue, subsequent works \cite{niu2022efficient, song2023ecotta, lee2024becotta} employ sample selection strategies to mitigate the harmful effects. More recent methods \cite{dobler2023robust, gan2023decorate, liu2023vida, liu2024continual, yu2024domain, zhang2024decomposing} leverage the mean teacher, obtained through historical model ensembling, to provide stable supervision.

Despite the promising performance along this path, existing methods still suffer from two problems: \textbf{1) Prediction adjustment, which adapts shallow layers based on the deep-layer outputs, leads to dilatory exploration of a new domain.} Domain shifts in low-level features should be corrected in the shallow layers of deep models. However, due to information compression in deeper layers \cite{saxe2019information}, adapting shallow layers solely from final predictions is inefficient;
\textbf{2) Exploitation of historical knowledge relies on a generalizable model.} 
According to the no-free-lunch theory \cite{wolpert1997no}, a single model inevitably forgets previously acquired information when exploring a new domain. How to exploit historical models to prevent catastrophic forgetting is still an open problem.

Inspired by the above issues, we propose a mean teacher framework called \textbf{\textit{BEE}}, which achieves a \textit{\textbf{B}alance between \textbf{E}xploration and \textbf{E}xploitation} for CTTA. 
To address challenge \textbf{1)}, we propose a \textit{\textbf{M}ulti-level \textbf{C}onsistency \textbf{R}egularization (\textbf{MCR})} loss. Rather than aligning the final predictions, it directly aligns the intermediate features. However, the feature space typically has much higher dimensionality than the prediction space, and the curse of dimensionality renders similarity measures ineffective.
To overcome this, we incorporate a set of codebooks to shrink the representation space and accelerate adaptation. 
Furthermore, to extend the scope of the exploration from the current batch to the entire current domain, MCR is applied to randomly selected recent samples before predicting the current batch.

In search of an answer to challenge \textbf{2)}, we employ a \textit{\textbf{C}omplementary \textbf{A}nchor \textbf{R}eplay (\textbf{CAR})} mechanism, which adaptively recovers forgotten knowledge from historical models.
Specifically, during the CTTA process, we periodically store the model to form an anchor pool. When a sudden spike in the MCR loss is detected, the current model lacks knowledge of the domain encountered. At this point, we select complementary models from the anchor pool and integrate them with the student model.

In a nutshell, the main contributions of this paper are summarized as follows:
\begin{itemize}
\item We propose MCR, a multi-level consistency regularization to align intermediate features between the teacher and student, accelerating the exploration of the current domain.
\item To exploit historical knowledge, we propose a CAR mechanism that reuses complementary anchors when facing significant domain shifts, thereby restoring the lost knowledge from the source and previous domains.
\item Empirical analyses on three CTTA benchmarks demonstrate that the proposed method achieves a better exploration-exploitation balance, and consistently outperforms existing CTTA methods.
\end{itemize}
\section{Related Work}
\begin{figure*}[htbp]
    \centering
    \includegraphics[width=\textwidth]{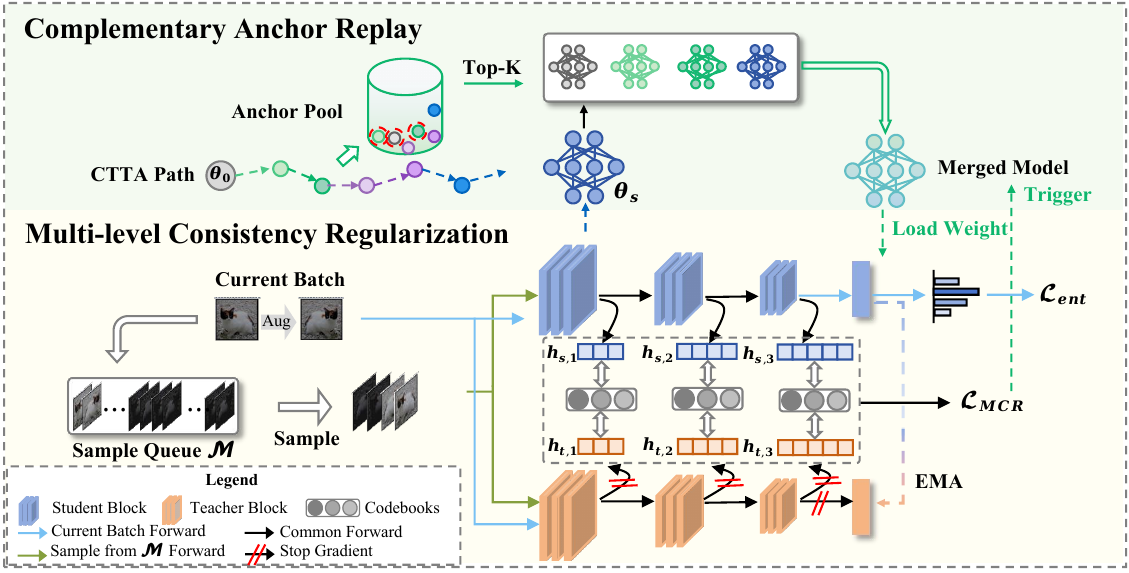}
    \caption{The architecture of our proposed framework. \textbf{First}, we introduce a \textit{Multi-level Consistency Regularization (\textbf{MCR})} loss for rapid exploration in the current domain. To extend exploration of the current domain, we apply MCR on randomly selected recent samples before predicting the current batch. \textbf{Second}, we employ a \textit{Complementary Anchor Replay (\textbf{CAR})} module that reuses previous checkpoints to handle a new domain. When a significant domain shift is detected by $\mathcal{L}_{MCR}$, the student model is reset by a weighted merge of selected anchors to recover forgotten knowledge.}
    \label{fig:framework}
\end{figure*}

In this section, we first present an overview of the existing research on the CTTA task in \cref{sec:related-work-ctta}. Next, we briefly introduce the background of the two approaches relevant to our proposed method in \cref{sec:related-work-sskd} and \cref{sec:related-work-model-merging}.

\subsection{Continual Test-time Adaptation} \label{sec:related-work-ctta}
Based on the adaptation objective, existing continual test-time adaptation methods can be broadly divided into three categories:

\textbf{1) BN calibration} methods focus on adjusting BN statistics (i.e., the mean and variance) to bridge the domain gap between the source and target domains, typically in a training-free manner. Early attempts \cite{nado2020evaluating} replaced the training BN statistics with those estimated on each test batch. Subsequent works refined this statistic estimation with various techniques \cite{you2021test, hong2023mecta, su2024unraveling, gong2022note, su2024towards}. 

\textbf{2) Entropy minimization} has emerged as a popular strategy for handling unlabeled data \cite{grandvalet2004semi, saito2019semi}. In the context of test-time adaptation, Tent \cite{wang2020tent} was the first to apply this approach by minimizing the mean entropy over test batches, thereby encouraging the model to produce confident predictions.
However, relying on noisy pseudo-labels can lead to severe error accumulation during continuous updates. To mitigate these harmful effects, later methods have employed sample selection strategies \cite{niu2022efficient, niu2023towards, song2023ecotta, lee2024becotta} based on the reliability of testing samples. 

\textbf{3) Consistency regularization} methods aim to enforce consistent network predictions under variations in input data or model parameters. 
Mainstream methods \cite{wang2022continual, chen2022contrastive, gan2023decorate, liu2023vida, yu2024domain, zhang2024decomposing, zhu2024reshaping} adopt a mean teacher framework \cite{tarvainen2017mean} to constrain the predictions of the student and teacher models to be consistent. 
For instance, CoTTA \cite{wang2022continual} employs multiple augmentations to refine the pseudo-labels generated by the teacher network, thereby guiding the student model toward more reliable predictions. 
Furthermore, RMT \cite{dobler2023robust} introduces a symmetric cross-entropy loss to mitigate abrupt gradient changes, making it a more robust consistency measure within the mean teacher framework. 
To preserve generalization, recent research has focused on updating the mean teacher framework using a limited number of trainable parameters via techniques such as prompt learning \cite{gan2023decorate, yang2024exploring}, adapters \cite{liu2023vida}, or block selection \cite{yu2024domain}.

Motivated by insights from the information bottleneck theory \cite{saxe2019information}, we propose a mean teacher framework that enforces multi-level consistency between the intermediate features of the student and teacher models, alleviating the impact of information compression in updating shallow parameters during adaptation.

\subsection{Self-supervised Knowledge Distillation}  \label{sec:related-work-sskd}
Self-supervised learning plays an important role in the absence of labeled data, with self-distillation forming the foundation of many leading works \cite{grill2020bootstrap, caron2021emerging, chen2021exploring, baevski2022data2vec}. In these approaches, knowledge is transferred from a teacher to a student model, which share an identical architecture.
A pioneering work, BYOL \cite{grill2020bootstrap}, introduces a framework where an online network and a target network update each other’s weights, demonstrating that self-supervision can work effectively without contrastive pairs. 
Building on this, SimSiam \cite{chen2021exploring} removes the momentum encoder from BYOL and employs a stop-gradient mechanism to prevent mode collapse, achieving effective knowledge distillation.
Moreover, DINO \cite{caron2021emerging} aligns student and teacher representations across different scales and augmentations using a distinct loss function and backbone, resulting in strong performance on various downstream tasks. 
To further enhance representation learning, MOKD \cite{song2023multi} integrates self-distillation and cross-distillation to improve semantic feature alignment between the two networks.
These methods have demonstrated strong performance in semi-supervised learning \cite{fini2023semi} and unsupervised domain adaptation \cite{zhang2023unsupervised}, indicating their ability to leverage unlabeled data and their potential for handling noisy pseudo-labels in CTTA. In contrast, the methods mentioned above rely on a pretrain-and-inference paradigm, whereas our setting involves continuous model updates during CTTA. For this specific constraint, we propose a strategy that simultaneously leverages knowledge from recent data and past anchors, balancing new information extraction and historical information reuse to achieve accurate predictions.

\subsection{Model Merging}  \label{sec:related-work-model-merging}
Model merging is an effective technique that combines parameters from multiple models to build a universal model, which preserves their original capabilities and even outperforms multi-task learning \cite{ainsworth2022git,li2022branch}. It also reduces the tendency of a single model to overfit specific samples or noise, thereby enhancing prediction accuracy, diversity, and robustness \cite{li2023deep}.
Among the early approaches, weight averaging \cite{wang2020federated,wortsman2022model,izmailov2018averaging} provides a direct and efficient way to fuse individual models, particularly when they are similar with certain differences in the weight space.
Task Arithmetic~\cite{ilharco2023editing} adopts a difference-based averaging strategy by adding the pre-trained model to a linear combination of task vectors, resulting in better performance in all tasks. 
Building on these ideas, later works consider the varying importance of individual models \cite{matena2022fishermerging, jang2024model,jin2023regmean} and conflicts between them \cite{yadav2024ties,marczak2024magmax} to determine interpolation coefficients. 

In this work, we propose a prediction discrepancy-based weighting scheme, prioritizing historical checkpoints most complementary to the current model to recover forgotten knowledge.
 
\section{Methodology} \label{sec:method}
\subsection{Overview}
\noindent\textbf{Task Definition.}\quad Given a pre-trained model $f_{\bm{\theta}}$ trained on source data $\mathcal{D}_S = (\mathcal{X}^S, \mathcal{Y}^S)$, our goal is to improve its performance during inference on a continually changing target domain. During testing, the model is evaluated on unlabeled target samples from several unknown target domains, denoted as $\mathcal{D}_{T_i} = \{(\mathcal{X}^{T_i})\}^{n_{\text{dom}}}_{i=1}$. In an online setting, target data arrives sequentially in batches, and the model must generate predictions for incoming batches in real time. Unlike traditional machine learning, CTTA allows adapting the model parameters at the inference time to bridge the generalization gap. Notably, previous predictions cannot be modified.

\noindent \textbf{Overall Framework.}\quad As illustrated in \cref{fig:framework}, our proposed method builds on a mean teacher framework \cite{yu2024domain} and comprises two core components: a \textit{\textbf{M}utli-level \textbf{C}onsistency \textbf{R}egularization (\textbf{MCR})} loss and a \textit{\textbf{C}omplementary \textbf{A}nchor \textbf{R}eplay (\textbf{CAR})} module. As described in \cref{sec:mcr}, to accelerate adaptation to the current domain, we enforce a multi-level consistency regularization across intermediate features. As shown in \cref{sec:car}, when the current model fails to handle the new domain, the proposed framework adaptively reuses previous checkpoints to recover forgotten knowledge.

\subsection{Multi-level Consistency Regularization} \label{sec:mcr}
\noindent\textbf{Preliminaries.} Before formally presenting our method, we briefly introduce the mean teacher framework. This framework contains two branches with identical architectures: a student model $f_{\bm{\theta}_s}$ and a teacher model $f_{\bm{\theta}_t}$. Taking the student model as an example, it can be decomposed into a series of blocks: $\{\Phi_{s,j}\}_{j=1}^{L}$, where $L$ is the number of blocks. Given an input image $\bm{x}$, we denote the intermediate features from the $j$-th block as:
\begin{equation}
\bm{h}_{s,j}=\Phi_{s,j}(\bm{h}_{s,j-1}) \in \mathbb{R}^{D_j}, ~~ \bm{h}_{s,0}=\bm{x}.
\end{equation}

Then the final output is rewritten as:
\begin{equation}
    f_{\bm{\theta}_s}(\bm{x}) = \Phi_{s,L} \circ \Phi_{s,L-1} \circ \cdots \circ \Phi_{s,1}(\bm{x}).
\end{equation}
For the teacher model, we denote $\Phi_{t,j}$ and $\bm{h}_{t,j}$ similarly.

Given a test batch $\{\bm{x}_i\}_{i=1}^N$, previous CTTA methods \cite{wang2022continual,yu2024domain} perform consistency regularization on the final predictions:
\begin{equation}
    \mathcal{L}_{con} = \frac{1}{N} \sum_{i=1}^N H(f_{\bm{\theta}_t}(\bm{x}_i), f_{\bm{\theta}_s}(\bm{x}’_i)),
\end{equation}
where $\bm{x}_i'$ is augmented from $\bm{x}_i$, $H(\bm{p},\bm{q}) = - \sum_{c=1}^C \bm{p}_c \cdot \log \bm{q}_c$ is the cross-entropy loss, and $C$ is the number of classes. The student $f_{\bm{\theta}_s}$ is updated by minimizing the consistency regularization loss, while teacher parameters are updated via the Exponential Moving Average (EMA) strategy:
\begin{equation} \label{eq:ema}
    \bm{\theta_t}\leftarrow m\,\bm{\theta}_t + (1 - m)\bm{\theta_s},
\end{equation}
where $m$ is the EMA momentum.

\noindent\textbf{Multi-level Loss}
As discussed in previous works \cite{yu2024domain,choi2022improving,zhou2022domain}, domain shifts typically affect the low-level statistics. Consequently, such shifts can be effectively mitigated through adaptation mechanisms applied to shallow blocks of the network. However, the above prediction consistency loss is inefficient in adapting the shallow layers due to the information compression of deep layers \cite{saxe2019information}. 

At first glance, a native solution is applying the consistency loss to intermediate features $\bm{h}_{s,j}$ and $\bm{h}_{t,j}$ directly. Unfortunately, in practice, such a simple method fails to resolve the issue empirically due to the following factors:
\begin{itemize}
    \item The dimensionality of intermediate features is much higher than the predictions, and the curse of dimensionality leads to the reduced discriminative capacity of the loss function;
    \item Only a few samples are available in the current batch, making the high-dimensional features sensitive to stochastic noise.
\end{itemize}

To address the above issues, we introduce a feature codebook \cite{van2017neural, caron2021emerging} to shrink the representation space of intermediate features. For all $j=1,\cdots,L$, we introduce $M$ representative vectors $\bm{q}_{s,j} \in \mathbb{R}^{D_j}$. Afterward, we replace the original features with the cosine similarities between $\bm{h}_{s,j}$ and $\bm{q}_{s,j}$ as the input to the loss function. Formally, let $\bm{Q}_{s,j} \in \mathbb{R}^{D_j \times M}$ denote a codebook in which each column is composed of $\bm{q}_{s,j}$, then the input of loss is as follows:
\begin{equation}
    \bm{p}_{s,j}(\bm{x}) = \sigma\left(\bm{Q}_{s,j}^\top \bm{h}_{s,j} / \tau_s\right) \in \mathbb{R}^{M},
\end{equation}
where $\sigma$ is the softmax function, $\tau_s > 0$ is a temperature hyperparameter controlling the sharpness of the feature distribution. All the intermediate features $\bm{h}_{s,j}$ and $\bm{q}_{s,j}$ are assumed to be L2-normalized. In this way, the representation space is compressed to the similarity with finite vectors, which is a technique commonly employed in representation sparsification for image generation \cite{van2017neural}. Similarly, we map the teacher outputs as follows:

\begin{equation}
  \bm{p}_{t,j}(\bm{x}) = \sigma\left(\bm{Q}_{t,j}^\top \bm{h}_{t,j} / \tau_t)\right).
\end{equation}
Unfortunately, we find that directly applying the same process leads to mode collapse, i.e., predicted distributions become overly sharp and nearly identical across inputs, undermining consistency. To avoid the trivial solution, following recent self-supervised learning work \cite{caron2020unsupervised,oquab2023dinov2}, we transfer the teacher outputs into a uniform distribution with a minimum transport cost. Concretely, let $\bm{P}_{t,j} \in \mathbb{R}^{N \times M}$ be a set of $\bm{p}_{t,j}(\bm{x}_i), i=1,\cdots,N$, we aim to transport $\bm{P}_{t,j}$ to $\widetilde{\bm{P}}_{t,j}$ according to the following objective:
\begin{equation}
\begin{aligned}
\max_{\widetilde{\bm{P}}_{t,j} \in \mathbb{R}_+^{N \times M}} &\langle \bm{P}_{t,j}, \widetilde{\bm{P}}_{t,j} \rangle \quad \\ 
\text{s.t.} \quad &\widetilde{\bm{P}}_{t,j}\bm{1}_{M} = \frac{1}{N}\bm{1}_{N}, \widetilde{\bm{P}}_{t,j}^\top\bm{1}_{N} = \frac{1}{M}\bm{1}_{M}.
\end{aligned}
\end{equation}
The constraint $\widetilde{\bm{P}}_{t,j}\bm{1}_{M} = \frac{1}{N}\bm{1}_{N}$ enforces a uniform distribution across the dimensions, thereby preventing mode collapse; The constraint $\widetilde{\bm{P}}_{t,j}^\top\bm{1}_{N} = \frac{1}{M}\bm{1}_{M}$ ensures that the output follows a probabilistic distribution, making it suitable for the cross-entropy loss. The optimization problem can be solved via the Sinkhorn-Knopp algorithm \cite{cuturi2013sinkhorn} (see Appendix), and here we abbreviate it as $\sigma_{SK}$:
\begin{equation}
    \widetilde{\bm{p}}_{t,j}(\bm{x}) = \sigma_{SK}\left(\bm{Q}_{t,j}^\top \bm{h}_{t,j} / \tau_t)\right).
\end{equation}

With the above techniques, we could apply the consistency regularization to intermediate features, leading to the final loss:
\begin{equation} \label{eq:mcr}
\begin{aligned}
    \mathcal{L}_{MCR} &= \sum_{j=1}^L \mathcal{L}_{MCR}^j~, \\
    \mathcal{L}_{MCR}^j &= \frac{1}{N} \sum_{i=1}^N H(\bm{p}_{s,j}(\bm{x_i}), \widetilde{\bm{p}}_{t,j}(\bm{x_i})).
\end{aligned}
\end{equation}

\begin{figure}[t]
    \centering
    \includegraphics[width=\linewidth]{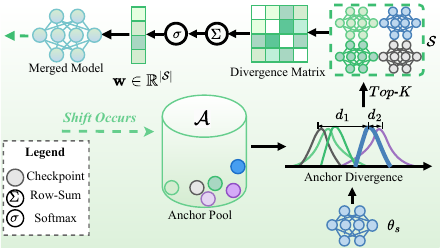}
    \caption{Structure of the CAR Module. When a sudden spike in MCR loss is detected, we update the student model with a super model obtained by merging the current student with the anchors of Top-$K$ largest prediction divergence.}
    \label{fig:enter-label}
\end{figure}

\begin{figure}[t]
    \centering
    \includegraphics[width=\linewidth]{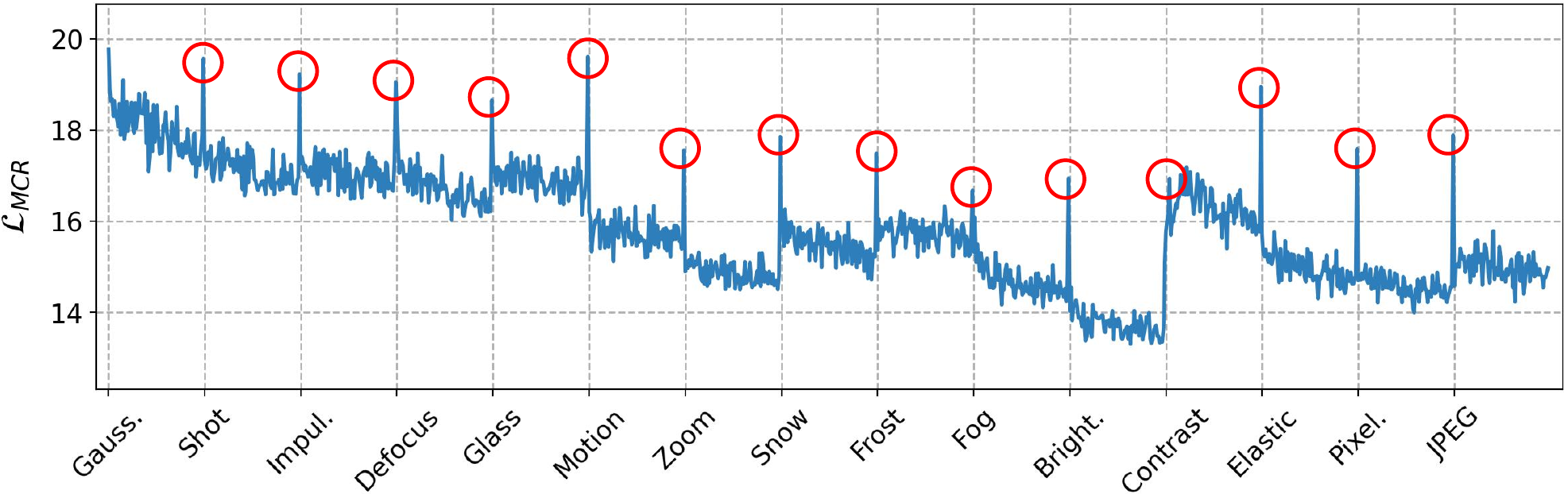}
    \caption{Plot of $\mathcal{L}_{\text{MCR}}$ over 15 sequential target domains. A sudden spike occurs in $\mathcal{L}_{\text{MCR}}$ when domain shifts, as highlighted with red circles.}
    \label{fig:aloss-trigger}
\end{figure}

\begin{algorithm}[tb]
  \caption{BEE Algorithm}
  \label{alg:bee}
  \LinesNumbered
  \KwIn{An incoming test batch $\{\bm{x}_i\}_{i=1}^N$, student model $f_{\bm{\theta}_s}$, teacher model $f_{\bm{\theta}_t}$, sample queue $\mathcal{M}$, inner steps $R$, inner batch size $B$, anchor pool $\mathcal{A}$, anchor storage period $\xi$, and time step $p$.}
  \KwOut{Predictions for the current batch.}
    Store the current batch in $\mathcal{M}$. \\
    \textcolor{lightblue}{$\triangleright$ Model update with recent samples}\\
    \For{$i = 1$ {\bf to} $R$}{
        Randomly sample a batch $\{\tilde{\bm{x}}_i\}_{i=1}^{B}$ from \(\mathcal{M}\). \\
        Update $\bm{\theta}_s$ with $\mathcal{L}_{MCR}$ (\cref{eq:mcr}). \\
        Update $\bm{\theta}_t$ with EMA (\cref{eq:ema}). \\
    }
    \textcolor{lightblue}{$\triangleright$ Model update with the current batch}\\
    Get model predictions with \cref{eq:prediction}. \\
    Update $\bm{\theta}_s$ with $\mathcal{L}_{MCR}$ in \cref{eq:mcr}. \\
    Update $\bm{\theta}_s$ with $\mathcal{L}_{ent}$ in \cref{eq:ent}. \\
    Update $\bm{\theta}_t$ with EMA in \cref{eq:ema}. \\
    \textcolor{lightblue}{$\triangleright$ Complementary anchor replay}\\
    \If {\(p \bmod \xi = 0\)}{
        Store student parameters in $\mathcal{A}$. 
    }
    Detect whether domain shifts with the z-score algorithm. \\
    \If{domain shift occurs}{
        \For{$k = 1$ {\bf to} $|\mathcal{A}|$}{
            Load parameters $\bm{\theta^{(k)}}$ from $\mathcal{A}$.  \\
            Calculate divergence with $d(\bm{\theta}^{(k)}, \bm{\theta}_s)$ in \cref{eq:sym-kl}. \\
        }
        Construct set $\mathcal{S}$ with $\bm{\theta}_s$ and selected anchors in \cref{eq:candidate-set}. \\
        Calculate the ensemble weight $\mathbf{w}$ in \cref{eq:ensemble-weight}. \\
        Update $\bm{\theta}_s$ via weighted model merging in \cref{eq:model-merging}. \\
    }
    \Return
\end{algorithm}

\subsection{Complementary Anchor Replay} \label{sec:car}
In this subsection, we focus on exploiting historical knowledge to mitigate performance degradation caused by overfitting. Existing methods \cite{wang2022continual, brahma2023probabilistic, press2024rdumb} propose resetting partial parameters to the pre-trained source model.
However, this sacrifices knowledge gained from previous test domains, which may include information valuable for the current domain. As mentioned in \cref{sec:intro}, a single model cannot capture all previously acquired information. To leverage previous knowledge more effectively, this work proposes to utilize multiple historical checkpoints by merging them into a super model for replay. Technically, we have to address the following problems:

\begin{itemize}
    \item Determine the timing of replay. Intuitively, replaying is necessary only when a significant domain shift occurs to handle the abrupt shift \cite{hoang2024persistent}, and an early replay might disrupt ongoing adaptation.
    \item Merge the historical checkpoints with adaptive weights. This work argues that knowledge complementary to the current model is essential for improving its generalization capability.
\end{itemize}

To overcome the first problem, we introduce a \(\mathcal{L}_{MCR}\)-based trigger: we monitor the multi‑level consistency regularization loss on each batch and reuse anchors as soon as it spikes sharply. As shown in \cref{fig:aloss-trigger}, such spikes empirically indicate a significant domain shift and reveal that the current model no longer matches the incoming distribution. We further pinpoint the exact shift moment using a z‑score detection algorithm \cite{ahmad2016real} (detailed in appendix).

For the second problem, we maintain an anchor pool $\mathcal{A} = \{\bm{\theta}^{(k)}\}_{k=1}^{C_a}$ with a capacity $C_a$  that adds a new anchor every $\xi$ batches to capture evolving student model states.

When a sudden increase in $\mathcal{L}_{MCR}$ is detected, the current model and its similar anchors cannot handle the current domain. Consequently, 
\textit{\textbf{anchors whose predictions diverge most from the current student are prioritized, since a larger divergence indicates that the anchor holds more complementary knowledge forgotten by the current model.}} 

To measure this divergence, we compute the symmetric KL divergence between any two models' predicted probability distributions on the current batch:
\begin{equation} \label{eq:sym-kl}
    d(\bm{\theta}_s^{(i)}, \bm{\theta}_s^{(j)}) 
= \frac{1}{2}
\left[
D_{KL}\bigl(P^{(i)}\|P^{(j)}\bigr)
\;+\;
D_{KL}\bigl(P^{(i)}\|P^{(j)}\bigr)
\right],
\end{equation}
where $P^{(i)} = \{\sigma(f_{\bm{\theta}_s^{(i)}}(\bm{x}_k))\}_{k=1}^{N}$ and $P^{(j)}$ is defined similarly.

Then we select a subset of anchors with the largest divergence from $\mathcal{A}$ for subsequent model ensembling:

\begin{equation} \label{eq:candidate-set}
    \mathcal{S} = \{\bm{\theta}_s\} \cup \{\bm{\theta}\,|\, d(\bm{\theta}_s, \bm{\theta}) \in \text{Top-}K\big(d(\bm{\theta}_s, \bm{\theta}')\big), \bm{\theta},\bm{\theta}'\in \mathcal{A}\}.
\end{equation}

Next, we compute symmetric KL divergences among all models in $\mathcal{S}$ and normalize the total divergence of each model $\bm{\theta} \in \mathcal{S}$ to derive the ensemble weight:
\begin{equation} \label{eq:ensemble-weight}
    \tilde{w}(\bm{\theta}) = \sum_{\bm{\theta}' \in \mathcal{S}} d(\bm{\theta}, \bm{\theta}'), ~ 
    w(\bm{\theta}) = \frac{\exp(\tilde{w}(\bm{\theta}))}
    {\sum_{\bm{\theta}' \in \mathcal{S}} \exp(\tilde{w}(\bm{\theta}'))}.
\end{equation}
Finally, we update the student model via weighted model merging: 
\begin{equation} \label{eq:model-merging}
    \bm{\theta}_s \leftarrow \sum_{\bm{\theta} \in \mathcal{S}} w(\bm{\theta}) \cdot \bm{\theta},
\end{equation}
thereby reintroducing historical information while adapting to the new domain.

\subsection{Algorithmic Procedure} \label{sec:overall-procedure}
Following previous works \cite{dobler2023robust, yu2024domain}, we apply a warm-up stage to initialize the teacher model. We also adopt the block selection process from \cite{yu2024domain} to identify the shallow blocks most relevant to domain shifts. During the warm-up phase, we update only the shallow blocks and the codebooks $\{\bm{Q}_{s,j}\}_{j=1}^L, \{\bm{Q}_{t,j}\}_{j=1}^L$ using 50,000 training samples. Afterward, the deep blocks and the codebooks are frozen during inference. 

At each time step $p$ in CTTA, given a testing batch $\{\bm{x}_i\}_{i=1}^N$, we first store it in a sample queue $\mathcal{M}$. To quickly explore the current domain, we randomly sample a batch $\{\tilde{\bm{x}}_i\}_{i=1}^{B}$ from $\mathcal{M}$ and update the student model with $\mathcal{L}_{MCR}$ for $R$ iterations. We obtain predictions for the current batch as follows:
\begin{equation} \label{eq:prediction}
\begin{aligned}
    \hat{Y} = \{\hat{y}_i\}_{i=1}^{N}, \quad
    \hat{y}_i = \frac{1}{2}\bigl(f_{\bm{\theta_s}}(\bm{x}_i) + f_{\bm{\theta_t}}(\bm{x}_i)\bigr).
\end{aligned}
\end{equation}
Subsequently, we update the student model with two backpropagation steps: first with $\mathcal{L}_{MCR}$, and second with the entropy minimization loss:
\begin{equation} \label{eq:ent}
    \mathcal{L}_{ent} = -\frac{1}{N}\sum_{i=1}^{N}\sum_{c=1}^C \hat{y}_{i,c} \log \hat{y}_{i,c}.
\end{equation}
Teacher parameters are updated using EMA. Finally, we trigger the CAR module once a sudden spike in the $\mathcal{L}_{MCR}$ is detected. The overall procedure is summarized in \cref{alg:bee}.
\section{Experiments}
\begin{table*}[t]
  \caption{Classification error rate ($\%$, lower is better) for the ImageNet-C benchmark using 
  ResNet-50. All results are evaluated at level 5. The best and second-best results are highlighted in \textbf{\textcolor{soft_red}{soft red}} and \textbf{\textcolor{soft_blue}{soft blue}}, respectively.}
  \label{tab:main-res-imagenet}
  \centering
  \resizebox{\textwidth}{!}{
  \begin{tabular}{l|l|ccccccccccccccc|c}
    \toprule
    \multicolumn{2}{c}{TIME} & \multicolumn{16}{c}{$t \rightarrow$} \\
    \midrule
    Backbone & Method & \rotatebox{80}{Gaussian} & \rotatebox{80}{Shot} & \rotatebox{80}{Impulse} & \rotatebox{80}{Defocus} & \rotatebox{80}{Glass} & \rotatebox{80}{Motion} & \rotatebox{80}{Zoom} & \rotatebox{80}{Snow} & \rotatebox{80}{Frost} & \rotatebox{80}{Fog} & \rotatebox{80}{Brightness} & \rotatebox{80}{Contrast} & \rotatebox{80}{Elastic} & \rotatebox{80}{Pixelate} & \rotatebox{80}{JPEG} & \rotatebox{80}{Mean} \\
    \midrule
    \multirow{11}[1]{*}{ResNet-50} 
    & Source only & 97.8 & 97.1 & 98.2 & 81.7 & 89.8 & 85.2 & 78.0 & 83.5 & 77.1 & 75.9 & 41.3 & 94.5 & 82.5 & 79.3 & 68.6 & 82.0\\
    & BN-1 \cite{nado2020evaluating} & 85.0 & 83.7 &85.0 &84.7 &84.3 &73.7 &61.2 &66.0 &68.2 &52.1 & \textbf{\textcolor{soft_blue}{34.9}} &82.7 &55.9 &51.3 &59.8 & 68.6\\
    & TENT-cont. \cite{wang2020tent} $_\text{ICLR'21}$ & 81.6 &74.6 &72.7 &77.6 &73.8 &65.5 & \textbf{\textcolor{soft_blue}{55.3} }&61.6 &63.0 &51.7 &38.2 &72.1 &50.8 &47.4 &53.3 & 62.6\\
    & AdaContrast \cite{chen2022contrastive} $_\text{CVPR'22}$ & 82.9 &80.9 &78.4 &81.4 &78.7 &72.9 &64.0 &63.5 &64.5 &53.5 &38.4 &66.7 &54.6 &49.4 &53.0 &65.6\\
    & CoTTA\cite {wang2022continual}$_\text{CVPR'22}$ & 84.7 &82.1 &80.6 &81.3 &79.0 &68.6 &57.5 &60.3 & 60.5 & 48.3 &36.6 &66.1 & \textbf{\textcolor{soft_blue}{47.2}} & \textbf{\textcolor{soft_red}{41.2}} & \textbf{\textcolor{soft_blue}{46.0}} & 62.7\\
    & EATA \cite{niu2022efficient} $_\text{ICML'22}$ &  75.9 & \textbf{\textcolor{soft_red}{65.7}} & \textbf{\textcolor{soft_red}{64.2}} & 74.6 & 69.7 & 64.7 & 57.6 & 62.8 & 63.0 & 53.5 & 40.8 & 64.5 & 50.2 & 47.8 & 50.0 & 60.3\\
    & SAR \cite{niu2023towards}$_\text{ICLR'23}$&  80.5 & 73.2 & 70.4 & 77.8 & 73.8 & 66.2 & 56.4 & 61.5 & 62.8 & 51.2 & 37.7 & 69.3 & 50.0 & 46.9 & 51.0 & 61.9\\
    & RMT \cite{dobler2023robust} $_\text{CVPR'23}$ &  77.9 & 73.1 &69.9 & \textbf{\textcolor{soft_blue}{73.5}} & \textbf{\textcolor{soft_red}{71.1}} & \textbf{\textcolor{soft_blue}{63.1}} & 57.1 & 57.1 & \textbf{\textcolor{soft_blue}{59.2} }&50.4 &42.9 & \textbf{\textcolor{soft_blue}{60.1}} & 49.0 &45.7 & 46.9 & 59.8\\
    & DPLOT \cite{yu2024domain} $_\text{CVPR'24}$ & \textbf{\textcolor{soft_red}{72.8}} & 70.4 & 69.4& \textbf{\textcolor{soft_red}{73.2}}& \textbf{\textcolor{soft_red}{71.1}}& 64.0&	58.1 & 58.5 & 60.1&	51.9& 45.4& 61.8	& 50.0& 47.2 &48.4	& 60.2 \\
    & OBAO \cite{zhu2024reshaping} $_\text{ECCV'24}$ & 78.5 & 75.3 & 73.0 & 75.7 & 73.1 & 64.5 & 56.0 & \textbf{\textcolor{soft_blue}{55.8}} & \textbf{\textcolor{soft_red}{58.1}} & \textbf{\textcolor{soft_blue}{47.6}} & 38.5 & \textbf{\textcolor{soft_red}{58.5}} & \textbf{\textcolor{soft_red}{46.1}} & \textbf{\textcolor{soft_blue}{42.0}} & \textbf{\textcolor{soft_red}{43.4}} & \textbf{\textcolor{soft_blue}{59.0}}\\
    & \textbf{BEE (ours)}& \textbf{\textcolor{soft_blue}{74.6}} & \textbf{\textcolor{soft_blue}{66.7}} & \textbf{\textcolor{soft_blue}{66.4}} & 74.0 & 73.2 & \textbf{\textcolor{soft_red}{61.4}} & \textbf{\textcolor{soft_red}{52.2}} & \textbf{\textcolor{soft_red}{55.5}} & 61.1 & \textbf{\textcolor{soft_red}{42.6}} & \textbf{\textcolor{soft_red}{32.6}} & 68.5 & 49.7 & 42.2 & 47.2 & \textbf{\textcolor{soft_red}{57.9}} \\
    \midrule
    \multirow{7}[1]{*}{ViT-base} 
    & Source Only & 53.0 & 51.8 & 52.1 & 68.5 & 78.8 & 58.5 & 63.3 & 49.9 & 54.2 & 57.7 & 26.4 & 91.4 & 57.5 & 38.0 & 36.2& 55.8 \\
    & Tent-cont \cite{wang2020tent} $_\text{ICLR'21}$ & 52.2 & 48.9 & 49.2 & 65.8 & 73.0 & 54.5 & 58.4 & 44.0 & 47.7 &  50.3 & 23.9 & 72.8 & 55.7 & 34.4 & 33.9 & 51.0 \\
    & CoTTA \cite{wang2022continual} $_\text{CVPR'22}$ & 52.9 & 51.6 & 51.4 & 68.3 & 78.1 & 57.1 & 62.0 & 48.2 & 52.7 & 55.3 & 25.9 & 90.0 & 56.4 & 36.4 & 35.2 & 54.8\\
    & VDP \cite{gan2023decorate} $_\text{AAAI'23}$ & 52.7 & 51.6 & 50.1 & 58.1 & 70.2 & 56.1 & 58.1 & 42.1 & 46.1 & 45.8 & 23.6 & 70.4 & 54.9 & 34.5 & 36.1 & 50.0 \\
    & ViDA \cite{liu2023vida} $_\text{ICLR'24}$ & 47.7 & 42.5 & 42.9 & 52.2 & 56.9 & 45.5 & 48.9 & 38.9 & 42.7 & \textbf{\textcolor{soft_blue}{40.7}} & 24.3 & \textbf{\textcolor{soft_blue}{52.8}} & 49.1 & 33.5 & 33.1 & 43.1\\
    & C-MAE \cite{liu2024continual} $_\text{CVPR'24}$ & \textbf{\textcolor{soft_blue}{46.3}} & \textbf{\textcolor{soft_blue}{41.9}} & \textbf{\textcolor{soft_blue}{42.5}} & \textbf{\textcolor{soft_blue}{51.4}} & \textbf{\textcolor{soft_blue}{54.9}} & \textbf{\textcolor{soft_blue}{43.3}} & \textbf{\textcolor{soft_red}{40.7}} & \textbf{\textcolor{soft_red}{34.2}} & \textbf{\textcolor{soft_red}{35.8}} & 64.3 & \textbf{\textcolor{soft_blue}{23.4}} & 60.3 & \textbf{\textcolor{soft_red}{37.5}} & \textbf{\textcolor{soft_red}{29.2}} & \textbf{\textcolor{soft_blue}{31.4}} & \textbf{\textcolor{soft_blue}{42.5}}\\
    & \textbf{BEE (ours)} & \textbf{\textcolor{soft_red}{41.6}} & \textbf{\textcolor{soft_red}{38.0}} & \textbf{\textcolor{soft_red}{37.8}} & \textbf{\textcolor{soft_red}{45.6}} & \textbf{\textcolor{soft_red}{48.6}} & \textbf{\textcolor{soft_red}{39.1}} & \textbf{\textcolor{soft_blue}{46.8}} & \textbf{\textcolor{soft_blue}{38.3}} & \textbf{\textcolor{soft_blue}{39.1}} & \textbf{\textcolor{soft_red}{35.6}} & \textbf{\textcolor{soft_red}{21.5}} & \textbf{\textcolor{soft_red}{47.9}} & \textbf{\textcolor{soft_blue}{41.8}} & \textbf{\textcolor{soft_blue}{29.6}} & \textbf{\textcolor{soft_red}{29.1}} & \textbf{\textcolor{soft_red}{{38.7}}} \\
    \bottomrule
  \end{tabular}}
\end{table*}

\subsection{Experimental Setup} \label{sec:exp-setup}
\noindent\textbf{Datasets and CTTA task setting.} We evaluate our method on three classification benchmarks under the CTTA setting, including CIFAR10-C, CIFAR100-C, and ImageNet-C. Each corruption dataset includes 15 corruption types (e.g., gaussian noise, shot noise, impulse noise, defocus blur) with 5 severity levels applied to the validation and test images of CIFAR and ImageNet, respectively. Following the CTTA setup in \cite{wang2022continual}, we adapt the source model sequentially to 15 target domains, each at the largest severity level 5. The entire adaptation process is carried out online, without indicating of when domain shifts occur. For each corruption type, we use 10,000 images from CIFAR10-C and CIFAR100-C, and 5,000 images from ImageNet-C.

\noindent \textbf{Baselines methods.} We compare our method with various state-of-the-art CTTA methods that categorized into three types: \textbf{1)} \textit{bn calibration method}, BN-1 \cite{nado2020evaluating}; \textbf{2)} \textit{entropy minimization methods}, including Tent-cont. \cite{wang2020tent}, EATA \cite{niu2022efficient}, and SAR \cite{liu2021towards}; \textbf{3)} \textit{consistency regularization methods}, including CoTTA \cite{wang2022continual}, AdaContrast \cite{chen2022contrastive}, RMT \cite{dobler2023robust}, DPLOT \cite{yu2024domain}, OBAO \cite{zhu2024reshaping}, VDP \cite{gan2023decorate}, ViDA \cite{liu2023vida}, and Continual-MAE \cite{liu2024continual}.

\noindent \textbf{Evaluation metric.} We report the classification error rate (\%) for each corruption domain, as well as the mean error across domains, following other baselines. 

\noindent \textbf{Implementation Details.} To maintain consistency with prior studies \cite{wang2022continual, yu2024domain, liu2023vida, liu2024continual}, we follow their implementation details. For experiments on CNN backbones, we adopt the pre-trained WideResNet-28-10 \cite{zagoruyko2016wide} model for CIFAR10-C, the pre-trained ResNeXt-29A \cite{xie2017aggregated} model for CIFAR100-C, and the pre-trained ResNet-50 \cite{he2016deep} model for ImageNet-C from Robustbench \cite{croce2020robustbench}. We set the batch size to 64 for ImageNet-C experiments and 200 for CIFAR datasets. For CNN backbones, we use an Adam optimizer with a learning rate of \(10^{-3}\); for ViT-base, we use a SGD optimizer with a learning rate of \(2\times 10^{-5}\). All CNN experiments are conducted on an NVIDIA RTX 4090 GPU, and ViT-Base experiments are run on a single NVIDIA A800 GPU. More details can be found in the appendix.

\begin{table}[t]
  \caption{Classification error rate ($\%$, lower is better) for the CIFAR10-C benchmark using WideResNet-28-10 and the CIFAR100-C benchmark using ResNeXt-29A. All results are evaluated at level 5. The best and second-best results are highlighted in \textbf{\textcolor{soft_red}{soft red}} and \textbf{\textcolor{soft_blue}{soft blue}}, respectively.}
  \label{tab:main-res-cifar-cnn}
  \centering
  \begin{tabular}{l|cc}
    \toprule
    Method & CIFAR10-C & CIFAR100-C \\
    \midrule
    Source only& 43.5 & 46.4 \\
    BN-1 \cite{nado2020evaluating}& 20.4 & 35.4 \\
    TENT-cont. \cite{wang2020tent} $_\text{ICLR'21}$ & 20.7 & 60.9 \\
    AdaContrast \cite{chen2022contrastive} $_\text{CVPR'22}$ & 18.5 & 33.4\\
    CoTTA  \cite{wang2022continual} $_\text{CVPR'22}$ & 16.2 & 32.5\\
    EATA \cite{niu2022efficient} $_\text{ICML'22}$ & 17.8 & 32.3\\
    SAR \cite{niu2023towards} $_\text{ICLR'23}$ & 20.3 & 31.9 \\
    RMT \cite{dobler2023robust} $_\text{CVPR'23}$ & 14.5 & 29.0\\
    DPLOT \cite{yu2024domain} $_\text{CVPR'24}$ & \textbf{\textcolor{soft_blue}{13.7}} & \textbf{\textcolor{soft_red}{27.8}} \\
    OBAO \cite{zhu2024reshaping} $_\text{ECCV'24}$ & 15.8 & 29.0 \\
    \textbf{BEE (ours)} & \textbf{\textcolor{soft_red}{12.7}} & \textbf{\textcolor{soft_red}{27.8}}\\
    \bottomrule
  \end{tabular}
\end{table}

\begin{table}[t]
    \caption{Classification error rate ($\%$, lower is better) for the CIFAR10-C and the CIFAR100-C benchmark using ViT-base. All results are evaluated at level 5. The best and second-best results are highlighted in \textbf{\textcolor{soft_red}{soft red}} and \textbf{\textcolor{soft_blue}{soft blue}}, respectively.}
    \label{tab:main-res-cifar-vit}
    \centering
    \resizebox{0.85\linewidth}{!}{
    \begin{tabular}{l|cc}
    \toprule
    Method & CIFAR10-C & CIFAR100-C\\
    \midrule
    Source Only & 28.2 & 35.4\\
    Tent-cont \cite{wang2020tent} $_\text{ICLR'21}$ & 23.5 & 32.1\\
    CoTTA \cite{wang2022continual} $_\text{CVPR'22}$ & 24.6 & 34.8\\
    VDP \cite{gan2023decorate} $_\text{AAAI'23}$ & 24.1 & 32.0 \\
    ViDA \cite{liu2023vida} $_\text{ICLR'24}$ & 20.7 & 27.3\\
    C-MAE \cite{liu2024continual} $_\text{CVPR'24}$ & \textbf{\textcolor{soft_blue}{12.6}} & \textbf{\textcolor{soft_blue}{26.4}}\\
    \textbf{BEE (ours)} & \textbf{\textcolor{soft_red}{11.8}} & \textbf{\textcolor{soft_red}{25.2}} \\
    \bottomrule
    \end{tabular}}
\end{table}

\begin{table}[t]
    \centering
    \caption{Ablation on components of BEE. Default settings are highlighted in \colorbox[rgb]{ .886,  .937,  .855}{green}. The best and second-best results are marked with \textbf{bold} and \underline{underline}.}
    \label{tab:ablation}
    \resizebox{\linewidth}{!}{
    \begin{tabular}{lccccccc|c}
        \toprule
        No. & $\mathcal{L}_{ent}$& $\mathcal{L}_{MCR}^4$ & $\mathcal{L}_{MCR}^3$ & $\mathcal{L}_{MCR}^2$ & $\mathcal{L}_{MCR}^1$ & $\mathcal{M} $& CAR & Mean Error$\downarrow$\\
        \midrule
        1 &\ding{51}& & & & & & &  70.2 \\
        2 & \ding{51}&\ding{51}& & & & & & 61.7 \\
        3 & \ding{51}&\ding{51}& \ding{51}& & & & & 59.2 \\
        4 & \ding{51}&\ding{51} & \ding{51}& \ding{51}& & & & 58.7\\
        5 & \ding{51}&\ding{51}& \ding{51} & \ding{51} & \ding{51}& & & 60.1\\
        6 & \ding{51}&\ding{51}& \ding{51} & \ding{51} & & \ding{51} & & 58.3\\
        7 & \ding{51}&\ding{51}& \ding{51} & \ding{51} & & & \ding{51} & 58.6\\
        8 & \cellcolor[rgb]{ .886,  .937,  .855} \ding{51}& \cellcolor[rgb]{ .886,  .937,  .855} \ding{51} & \cellcolor[rgb]{ .886,  .937,  .855}\ding{51} & \cellcolor[rgb]{ .886,  .937,  .855}\ding{51} & \cellcolor[rgb]{ .886,  .937,  .855} & \cellcolor[rgb]{ .886,  .937,  .855} \ding{51} & \cellcolor[rgb]{ .886,  .937,  .855} \ding{51} &\cellcolor[rgb]{ .886,  .937,  .855}57.9\\
        \bottomrule
    \end{tabular}}
\end{table}

\begin{table}[t]
    \centering
    \caption{Ablation on the anchor replay strategy of CAR. Default settings are highlighted in \colorbox[rgb]{ .886,  .937,  .855}{green}. The best and second-best results are marked with \textbf{bold} and \underline{underline}.}
    \label{tab:ablation-car}
    \resizebox{\linewidth}{!}{
    \begin{tabular}{lccccc|c}
        \toprule
        No. & \makecell{$\mathcal{L}_{MCR}$-based \\ trigger} & \makecell{Fix\\Interval}  & $\bm{\theta_0}$ & Averaging & \makecell{Weighted\\ Merging} & Mean Error$\downarrow$\\
        \midrule
        1 & & & & & & 58.3 \\
        2 & & 40 & & &\ding{51}& 59.4 \\
        3 & & 80 & & &\ding{51}& 58.4 \\
        4 & & 160 & & &\ding{51}& 58.2 \\
        5 & & 320 & & &\ding{51}& 58.0 \\
        6 & & 640 & & &\ding{51}& 58.1 \\
        7 & \cellcolor[rgb]{ .886,  .937,  .855}\ding{51}& \cellcolor[rgb]{ .886,  .937,  .855}& \cellcolor[rgb]{ .886,  .937,  .855}& \cellcolor[rgb]{ .886,  .937,  .855}&\cellcolor[rgb]{ .886,  .937,  .855}\ding{51}& \cellcolor[rgb]{ .886,  .937,  .855}\textbf{57.9}\\
        8 & \ding{51}& & \ding{51}& & & \underline{58.2} \\
        9 & \ding{51}& & &\ding{51}& & \underline{58.2} \\
        \bottomrule
    \end{tabular}}
\end{table}

\noindent \subsection{Main Results} \label{sec:main-res}
The evaluation results on three CTTA benchmarks with six different backbones are reported in \cref{tab:main-res-imagenet}, \cref{tab:main-res-cifar-cnn} and \cref{tab:main-res-cifar-vit}. We draw the following conclusions: \textbf{1)} Our proposed BEE consistently outperforms all baseline methods across datasets and model architectures. \textbf{2)} Compared with existing entropy minimization methods (e.g., Tent-cont., EATA, SAR), the proposed BEE improves performance by at least \textbf{2.4\%}, \textbf{5.1\%}, and \textbf{4.1\%} with CNN backbones on ImageNet-C, CIFAR10-C, and CIFAR100-C, respectively, and by \textbf{6.9\% $\sim$ 12.3\%} with ViT-base. As shown in the detailed results in \cref{tab:main-res-imagenet}, although these methods perform better in the first several domains, significant performance degradation occurs in later domains. These outcomes highlight the necessity of CAR in exploiting historical knowledge and preventing forgetting. \textbf{3)} Compared to advanced consistency regularization approaches (e.g., CoTTA, RMT, DPLOT, OBAO), BEE gets improvements of at least \textbf{1.1\%}, and \textbf{1.0\%} on ImageNet-C and CIFAR100-C using CNN backbones, while its performance is comparable to DPLOT on CIFAR100-C. With ViT-base, we still observe consistent improvements on three datasets. Moreover, BEE exhibits substantial performance gains in the early stages, thereby demonstrating the effectiveness of the MCR loss in accelerating exploration.

\subsection{Ablation Study} \label{sec:ablation}
To validate the effectiveness of each component in the proposed BEE, we conduct ablation experiments on ImageNet-C using the ResNet-50 backbone. The evaluation results are reported in \cref{tab:ablation} and \cref{tab:ablation-car}.

\noindent \textbf{Ablation results on BEE.} 
The comparison between \textit{line 1} and \textit{lines 2-5} in \cref{tab:ablation} (a) shows that, when applying the multi-level consistency regularization loss from deeper to shallower blocks ($\mathcal{L}_{MCR}^4 \rightarrow \mathcal{L}_{MCR}^1 $), the mean error first decreases and then increases. The optimal configuration is achieved by summing the losses from the last three blocks (\textit{line 4}), yielding an \textbf{11.5\%} performance improvement compared to using only $\mathcal{L}_{ent}$. Furthermore, comparing \textit{line 4} and \textit{line 6} shows that incorporating the sample queue $\mathcal{M}$ yields an additional \textbf{0.4\% }improvement. Considering the impact of CAR, comparing \textit{line 7} with \textit{line 4} and \textit{line 8} with \textit{line 6} reveals that adding CAR further reduces the mean error by \textbf{0.1\%} and \textbf{0.4\%}, respectively. Moreover, when the sample queue is used to accelerate adaptation to the entire current domain, the effect of CAR becomes even more significant.

\noindent \textbf{Ablation results on CAR.} A simple baseline is to reuse past anchors at fixed intervals. However, comparing \textit{line 7} with \textit{lines 2-6} in \cref{tab:ablation} (b) shows that our method achieves a \textbf{0.4\%} improvement, while fixed interval settings lead to significant performance drops. Since there is no prior knowledge of the optimal interval, dynamic review via the $\mathcal{L}_{MCR}$-based trigger is necessary. Furthermore, comparing \textit{line 7} with \textit{lines 8-9} reveals that simply using the source pre-trained model or applying basic anchor averaging yields only a 0.1\% improvement each. This result highlights the effectiveness of incorporating previously encountered domain knowledge and accounting for knowledge diversity.

\begin{figure}
     \centering
     \includegraphics[width=0.9\linewidth]{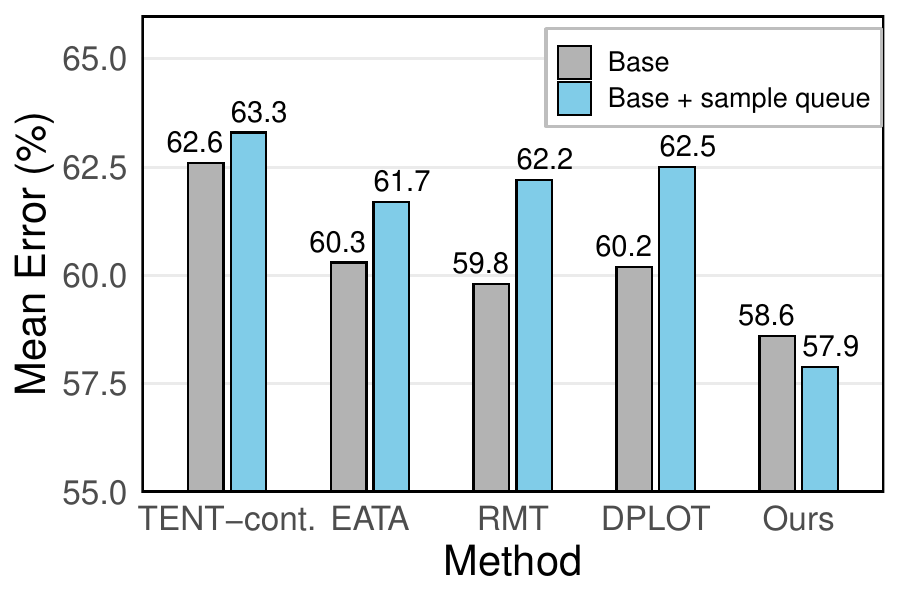}
     \caption{Impact of using the sample queue on baselines.}
     \label{fig:baseline-queue}
\end{figure}

\noindent \textbf{Impact of using the sample queue on other baselines.} As described in \cref{sec:overall-procedure}, we use a sample queue to draw recent batches for multi-level consistency regularization. To check if the performance gains arise from our objective or simply from extra backpropagation, we also add a sample queue to other baseline methods. We use the same sampling batch size and iteration, and re-tune the hyperparameters for each baseline. The results are presented in \cref{fig:baseline-queue}. All baseline methods show a performance drop after adding the sample queue, which indicates that simply increasing backpropagation iterations does not improve performance. This result supports the effectiveness of our objective design.

\begin{figure}
    \centering
    \includegraphics[width=0.9\linewidth]{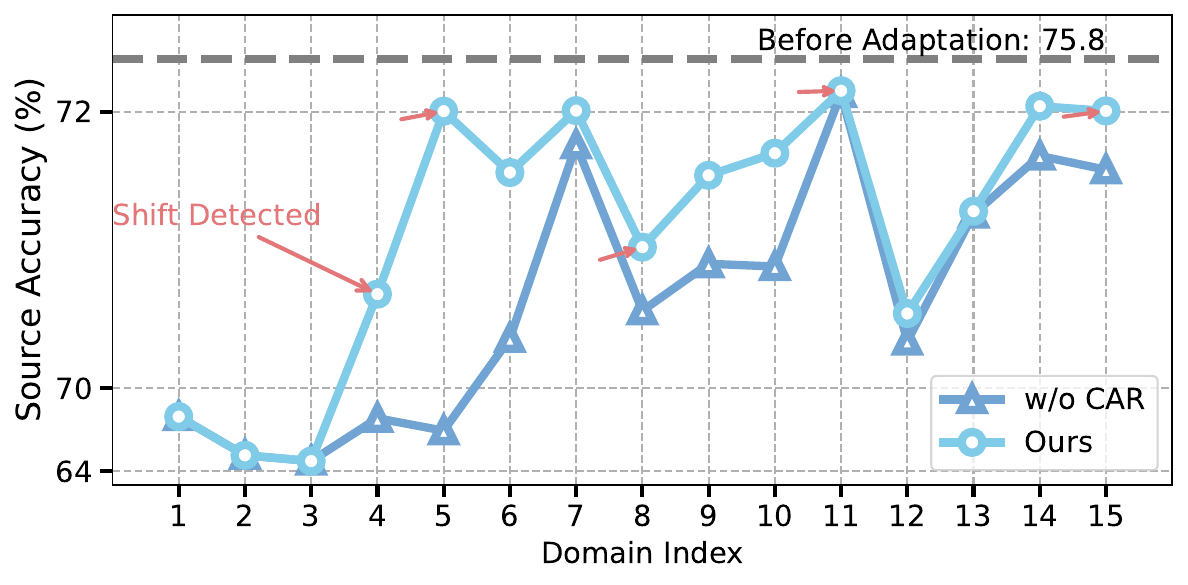}
    \caption{Comparison of results for our BEE with and without the CAR module on the source validation set.}
    \label{fig:antiforgetting-source}
\end{figure}

\noindent \textbf{Anti-forgetting capability evaluation.} In \cref{fig:antiforgetting-source}, we evaluate the effect of CAR on resisting forgetting. We store a checkpoint at the end of each domain and evaluate it on 5000 randomly sampled images from the ImageNet validation set. From the results, we observe the following: 1) A significant domain shift is closely associated with the forgetting of source knowledge. For instance, when the domain shifts from blur corruptions to weather corruptions ($7 \rightarrow 8$), source accuracy drops significantly; 2) CAR detects these shift moments and recalls source knowledge by reusing historical anchors, which boosts source accuracy or at least slows its decline.


\noindent \subsection{Hyperparameter Analysis} \label{sec:sensitivity}
Our proposed framework involves four hyperparameters: $R$ for the inner steps sampling recent batches, and $K$, $\xi$, and $C_a$ for CAR. Next, we analyze the effects of each parameter individually. All experiments are conducted on ImageNet-C with ResNet-50.

\begin{figure}
    \centering
    \begin{subfigure}[b]{0.48\linewidth}
        \centering
        \includegraphics[width=\linewidth]{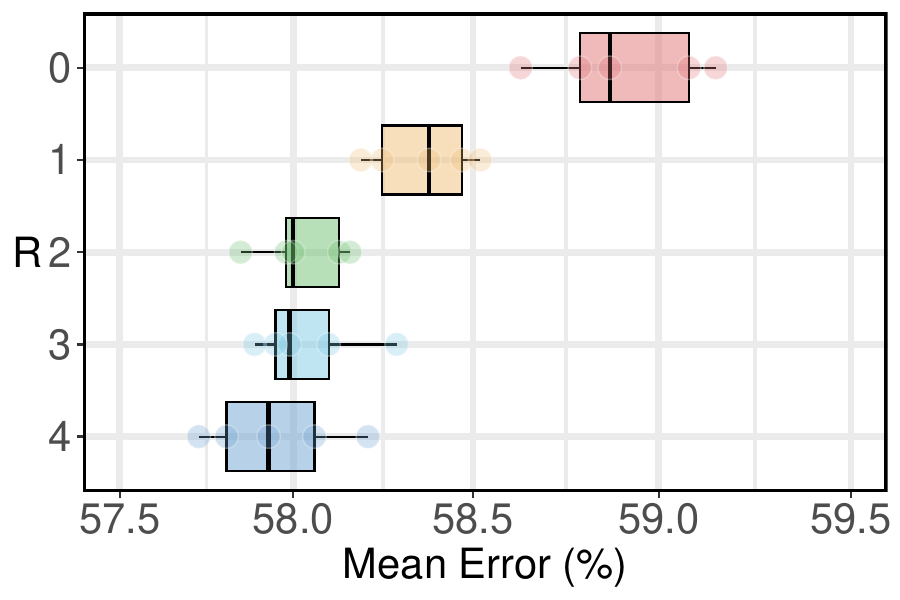}
        \vspace{-5mm}
        \caption{}
    \end{subfigure}
    \begin{subfigure}[b]{0.48\linewidth}
        \centering
        \includegraphics[width=\linewidth]{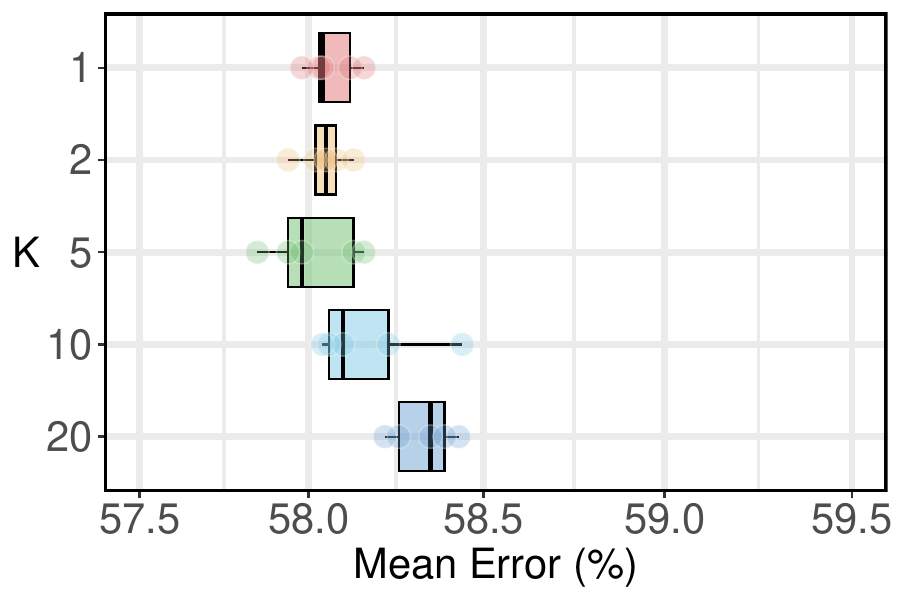}
        \vspace{-5mm}
        \caption{}
    \end{subfigure}
    \begin{subfigure}[b]{0.48\linewidth}
        \centering
        \includegraphics[width=\linewidth]{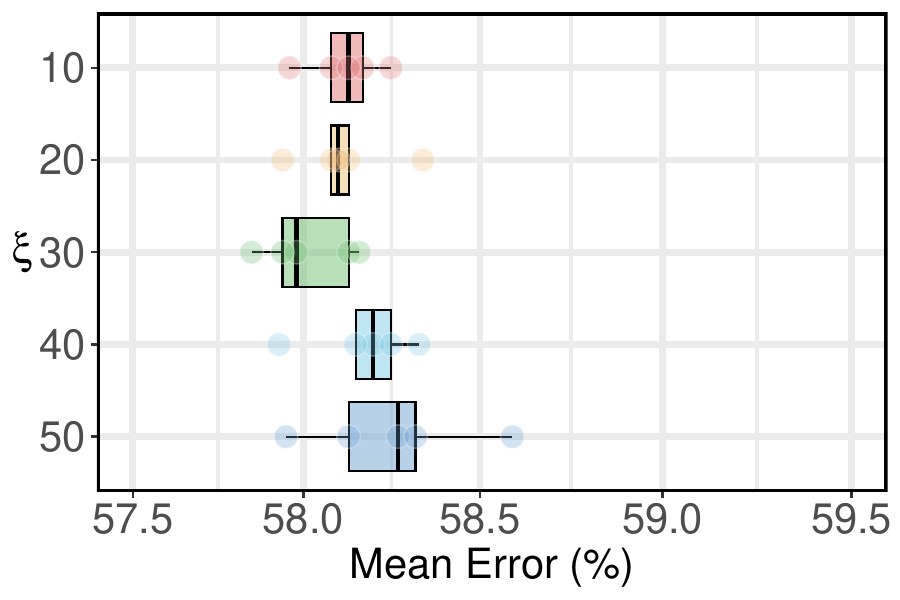}
        \vspace{-5mm}
        \caption{}
    \end{subfigure}
    \begin{subfigure}[b]{0.48\linewidth}
        \centering
        \includegraphics[width=\linewidth]{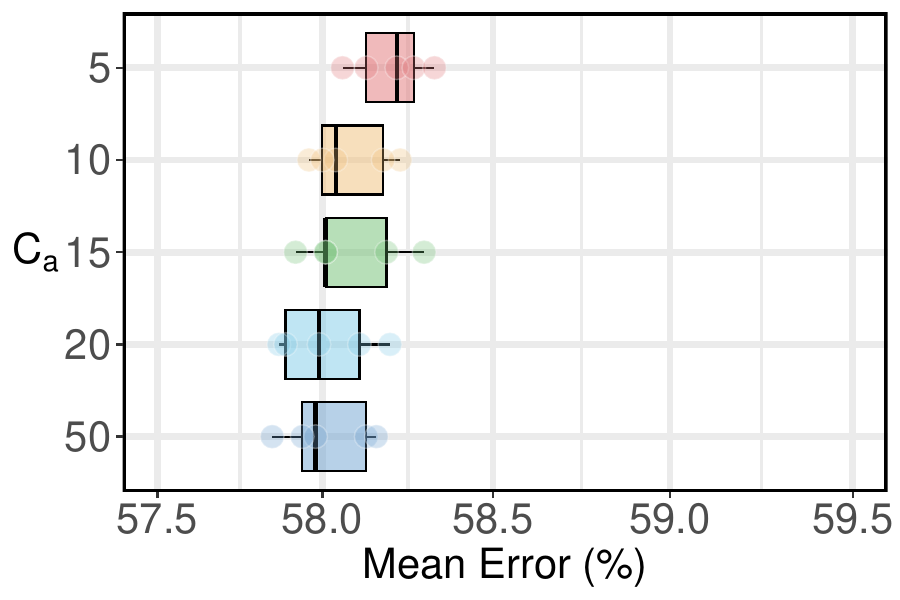}
        \vspace{-5mm}
        \caption{}
    \end{subfigure}
    \vspace{-3mm}
    \caption{Hyperparameter sensitivity analysis. (a) Sensitivity to inner steps $R$. (b) Sensitivity to Top-$K$. (c) Sensitivity to anchor storage period $\xi$. (d) Sensitivity to the anchor pool size $C_a$.}
    \label{fig:sensitivity}
\end{figure}

\noindent \textbf{Effect of $R$.} From \cref{fig:sensitivity} (a), we observe that as $R$ increases, the mean error continuously decreases, indicating that incorporating recent batches benefits exploration. However, the improvement slows down when $R > 2$. To balance efficiency and performance, we choose $R = 2$ for our main experiments.

\noindent \textbf{Effect of $K$.} As illustrated in \cref{fig:sensitivity} (b), introducing historical anchors consistently improves performance. The proposed method achieves the best performance when the Top-$K$ ratio $K$ is set to 5. As $K$ increases, performance declines due to redundant knowledge reducing diversity, which weakens the effectiveness of the merged model.

\noindent \textbf{Effect of $\xi$ and $C_a$.} As shown in \cref{fig:sensitivity} (c), increasing $\xi$ leads to a continuous decrease in mean error. However, if anchors are saved too infrequently, too few anchors will be available during early adaptation. Therefore, we set $\xi=30$. \cref{fig:sensitivity} (d) demonstrates that mean error decreases as $C_a$ increases. Since the memory overhead per anchor is minimal, we choose $C_a=50$. 

\section{Conclusion}
In this paper, we propose a mean teacher framework, which makes an appropriate exploration-exploitation balance for continual test-time adaptation. Within this framework, we introduce a multi-level consistency regularization loss to align intermediate features between the student and teacher, thereby accelerating the exploration in the current domain. In the face of diverse domain shifts, we design a complementary anchor replay mechanism to exploit historical information to recover forgotten knowledge. Experimental results demonstrate that our framework consistently outperforms competitive methods across different backbones and benchmarks. 
 
%
\bibliographystyle{IEEEtran}
\bibliography{main}

\clearpage
\section*{Appendix}
\startcontents[appendices]
\printcontents[appendices]{l}{1}{\setcounter{tocdepth}{2}}
\section{Additional Illustration of Method}

\subsection{Notations}
For convenience and clarity, we present a comprehensive list of the notations employed in our proposed method in \cref{tab:notations-all}.

\subsection{Details of Sinkhorn-Knopp algorithm}
As mentioned in \cref{sec:mcr}, we employ a Sinkhorn-Knopp algorithm to solve the optimal transport problem defined on the teacher intermediate features. Since the current batch carries only limited and noisy information, we aggregate teacher intermediate features from past and current batches using a feature queue of size 2048. We then enforce a uniform distribution constraint along both the feature queue and codebook dimensions. This process is repeated \(3\) times in our experiments. Below, we present the PyTorch-style pseudo-code for the Sinkhorn-Knopp algorithm.


    
\begin{lstlisting}[style=py]
def sinkhorn_knopp_teacher(p, temp, max_iters=3):
    p = p.float()
    new_p = torch.exp(p / temp)
    sum_p = torch.sum(new_p)
    new_p /= sum_p  
    # M: number of prototypes
    # N: number of features to assign
    M, N = new_p.shape
    
    for _ in range(max_iters):
        # Row normalization: 
        # total weight per prototype must be 1/M
        new_p /= torch.sum(new_p,dim=1,keepdim=True)
        new_p /= M
        # Column normalization: 
        # total weight per sample must be 1/N
        new_p /= torch.sum(new_p,dim=0,keepdim=True)
        new_p /= N
    
    # Make the columns sum to 1
    new_p *= N
    return new_p
\end{lstlisting}

\subsection{Details of the z-score algorithm }
In \cref{sec:car}, we apply a $\mathcal{L}_{MCR}$-based trigger to monitor significant domain shifts. Specifically, we maintain a sliding window to store loss values and compute short-term statistics. For each time step $p$, we calculate the z-score as follows:
\begin{equation}
    z = \frac{\mathcal{L}_{MCR}-\mu}{\sigma},
\end{equation}
where $\mu$ and $\sigma$ denote the mean and standard deviation of the values in the window. If $z$ exceeds the threshold $\eta$, we conclude that a domain shift has occurred and clear the sliding window. To maintain stability, we compute the z-score using the exponential moving average of $\mathcal{L}_{MCR}$ with a momentum of 0.9. In our experiments, we set the window size to 100 and $\eta = 1.5$.
\section{Detailed Description of Experiments}
\subsection{Implementation Details}
We follow the same warm-up setting as in \cite{dobler2023robust}, using 50,000 training samples and the same learning rate and batch size as in CTTA inference. We determine the range of shallow layer parameters most relevant to domain shifts via block selection \cite{yu2024domain}, and the results are shown in \cref{tab:shallow-layers}. During CTTA inference, we only update the selected shallow layer parameters.

As for the concept of a block mentioned in \cref{sec:method}, we follow the backbone's original parameter naming scheme. For ViT-base, we group every four blocks and treat them as a single unit for multi-level consistency regularization. Overall, block partitioning is detailed in \cref{tab:block-partioning}.

\begin{table}[tb]
\caption{Notations and descriptions used in BEE.}
\label{tab:notations-all}
\resizebox{\linewidth}{!}{
\begin{tabular}{l|c}
\toprule
Notations & Description \\
\midrule
\rowcolor[rgb]{0.8, 0.8, 0.8} \multicolumn{2}{c}{Model notations} \\
\midrule
$f_{\bm{\theta}}$ & Model with parameters $\bm{\theta}$ \\
$m$ & EMA update momentum \\
$L$ & Number of CNN or Transformer blocks \\
$D_j$ & Intermediate feature dimension of the $j$-th block \\
$M$ & Feature distribution dimension \\
$\Phi_{s,j}$ & The $j$-th block of the student model \\
$\Phi_{t,j}$ & The $j$-th block of the teacher model \\
$\bm{h}_{s, j}$ & Outputs of the $j$-th block of the student model \\
$\bm{h}_{t, j}$ & Outputs of the $j$-th block of the teacher model \\
$\bm{Q}_{s, j}$ & Codebook of the $j$-th block of the student model \\
$\bm{Q}_{t, j}$ & Codebook of the $j$-th block of the teacher model \\
\midrule
\rowcolor[rgb]{0.8, 0.8, 0.8} \multicolumn{2}{c}{MCR notations} \\
\midrule
$\mathcal{M}$ & Sample queue \\
$R$ & Iteratiions to sample recent batches from the sample queue \\
$B$ & Batch size of recent batches  \\
$\tau_{s}$ & \makecell[c]{Temperature controlling student distribution sharpness} \\
$\tau_{t}$ & \makecell[c]{Temperature controlling teacher distribution sharpness} \\
\midrule
\rowcolor[rgb]{0.8, 0.8, 0.8} \multicolumn{2}{c}{CAR notations} \\
\midrule
$\mathcal{A}$ & Anchor pool \\
$C_a$ & Anchor pool capacity \\ 
$K$ & Top-$K$ ratio to select anchors with most complementary knowledge \\
$\mathcal{S}$ & \makecell[c]{Anchor merging candidate set} \\
$\xi$ & Anchor update period \\ 
\bottomrule
\end{tabular}}
\end{table}

\begin{table}[b]
\caption{Shallow layer parameters updated during CTTA.}
\label{tab:shallow-layers}
\resizebox{\linewidth}{!}{
\begin{tabular}{l|ccc}
\toprule
Backbone & CIFAR10-C & CIFAR100-C & ImageNet-C \\
\midrule
WideResNet-28-10 & [block1.layer.0-3] & - & -\\
ResNeXt-29A & - & [stage1.0-2]& -\\
ResNet-50 & - & - & [layer1.0-2] \\
ViT-base & [blocks.0-5] &  [blocks.0-5]  &  [blocks.0-5] \\
\bottomrule
\end{tabular}}
\end{table}

\begin{table}[b]
\caption{Block Partitioning for Different Backbones.}
\label{tab:block-partioning}
\resizebox{\linewidth}{!}{
\begin{tabular}{l|cccc}
    \toprule
    Backbone & 1 & 2 & 3 & 4 \\
    \midrule
    WideResNet-28-10 & block1 & block2 & block3 & -\\
    ResNeXt-29A & stage1 & stage 2 & stage 3 & -\\
    ResNet-50 & layer1 & layer2 & layer3 & layer4 \\
    ViT-base & [blocks.0-3] &  [blocks.4-7]  &  [blocks.8-11] & - \\
    \bottomrule
\end{tabular}}
\end{table}
\begin{figure}[t]
    \centering
    \begin{subfigure}[b]{0.48\linewidth}
        \centering
        \includegraphics[width=\linewidth]{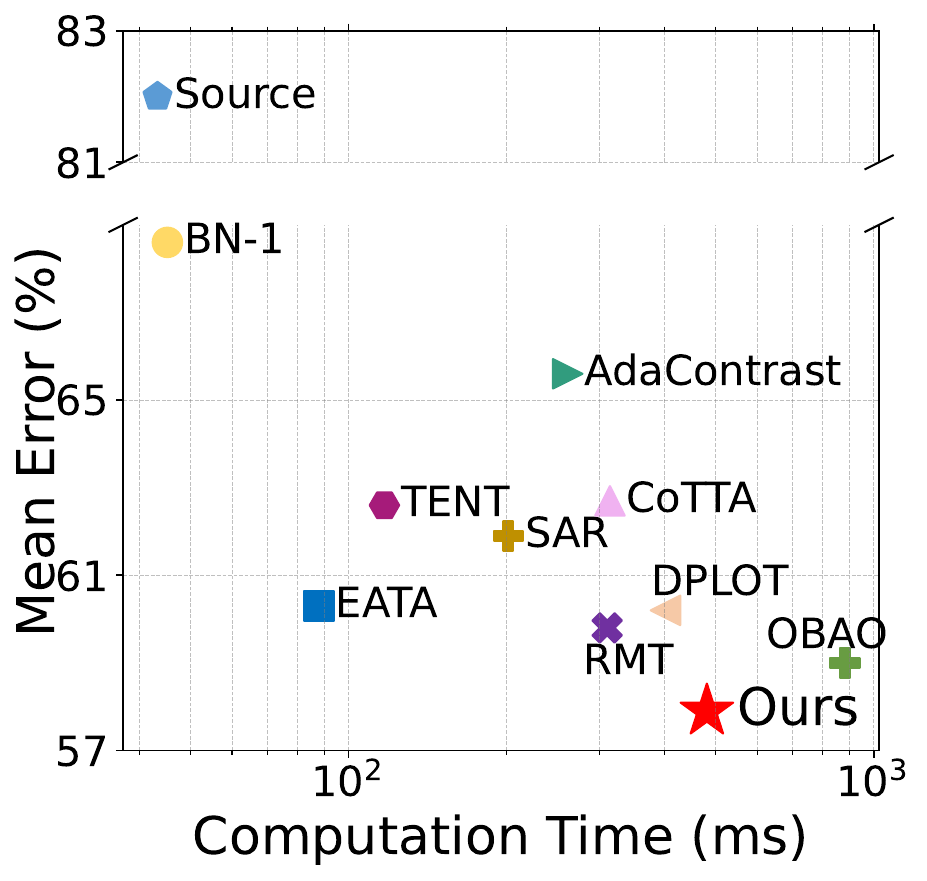}
    \end{subfigure}
    \begin{subfigure}[b]{0.48\linewidth}
        \centering
        \includegraphics[width=\linewidth]{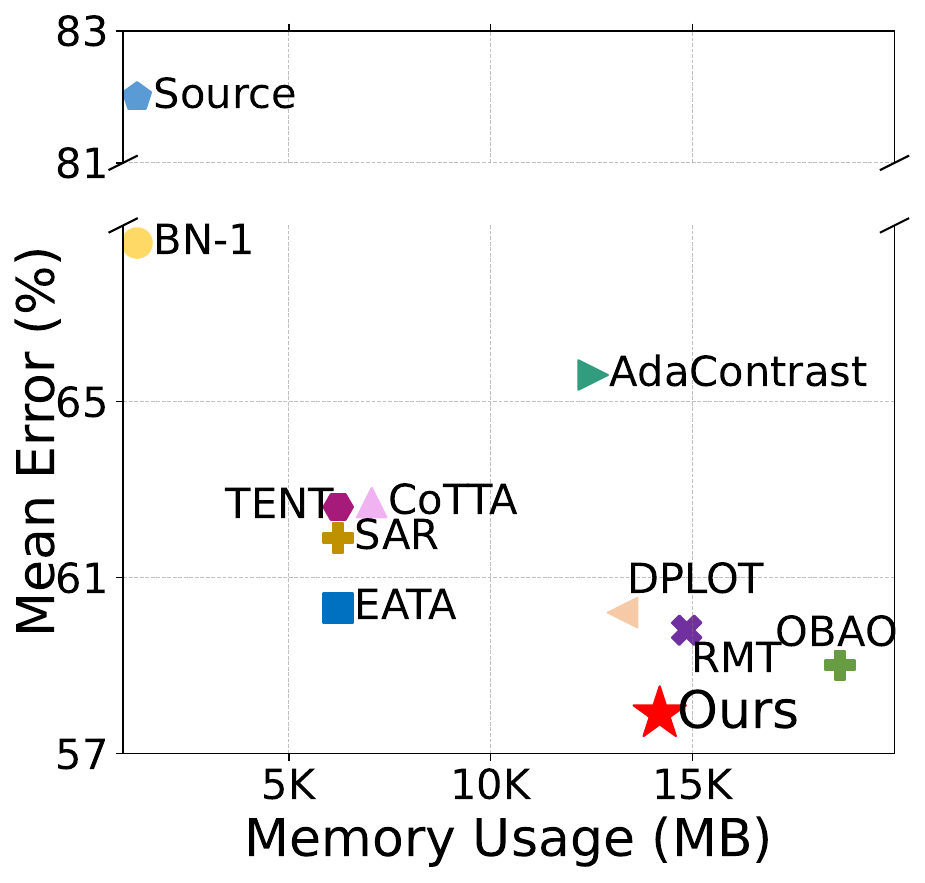}
    \end{subfigure}
    \caption{Analysis of time and memory usage of our proposed BEE and previous SOTA baselines on ImageNet-C with ResNet-50. Better performance is shown by results further toward the lower left.}
    \label{fig:efficiency}
\end{figure}

\subsection{Complexity Analysis} \label{sec:efficiency}
\noindent\textbf{Time and Memory Usage.} We report the evaluation results of adaptation speed (milliseconds per batch) and GPU memory usage (MB) in \cref{fig:efficiency}. The proposed BEE significantly outperforms the top baseline, improving mean error by 1.1 \% while using nearly half the computation time and \(3/4\) of the memory. Compared with other consistency regularization methods, our approach achieves an average improvement of 2.525\% while requiring only an extra 4.3 ms per batch and 729.3 MB of GPU memory.  Overall, BEE offers substantially superior adaptation performance with a manageable increase in memory and time consumption.

\vfill

\end{document}